\documentclass[11pt]{article}
\usepackage{acl2016}
\usepackage{times}
\usepackage{latexsym}
\usepackage{amsmath}
\usepackage{amssymb}
\usepackage{graphicx}
\usepackage{ifthen}
\usepackage[utf8]{inputenc}

\usepackage{algorithmicx}
\usepackage{algpseudocode}
\usepackage{color}
\usepackage{framed}
\usepackage[T1]{fontenc}
\usepackage{url}
\usepackage{xspace}

\newcommand\inner[2]{\ensuremath{\left< #1, #2 \right>}} 

\ifthenelse{\isundefined{\definition}}{\newtheorem{definition}{Definition}}{}
\ifthenelse{\isundefined{\assumption}}{\newtheorem{assumption}{Assumption}}{}
\ifthenelse{\isundefined{\hypothesis}}{\newtheorem{hypothesis}{Hypothesis}}{}
\ifthenelse{\isundefined{\proposition}}{\newtheorem{proposition}{Proposition}}{}
\ifthenelse{\isundefined{\theorem}}{\newtheorem{theorem}{Theorem}}{}
\ifthenelse{\isundefined{\lemma}}{\newtheorem{lemma}{Lemma}}{}
\ifthenelse{\isundefined{\corollary}}{\newtheorem{corollary}{Corollary}}{}
\ifthenelse{\isundefined{\alg}}{\newtheorem{alg}{Algorithm}}{}
\ifthenelse{\isundefined{\example}}{\newtheorem{example}{Example}}{}

\newcommand\citep\cite
\newcommand\citet\newcite

\newcommand{\atis}{\textsc{ATIS}\xspace}
\newcommand{\geo}{\textsc{Geo}\xspace}
\newcommand{\overnight}{\textsc{Overnight}\xspace}

\newcommand{\catroot}{\textsc{Root}\xspace}
\newcommand{\catsent}{\textsc{Sent}\xspace}

\newcommand{\catstate}{\textsc{State}\xspace}
\newcommand{\catstateid}{\textsc{StateId}\xspace}

\newcommand{\absent}{\textsc{AbsEntities}\xspace}
\newcommand{\absphrase}{\textsc{AbsWholePhrases}\xspace}
\newcommand{\concat}[1]{\textsc{Concat-#1}\xspace}

\newcommand{\vocabin}{\mathcal{V}^{\text{(in)}}}
\newcommand{\phiin}{\phi^{\text{(in)}}}
\newcommand{\vocabout}{\mathcal{V}^{\text{(out)}}}
\newcommand{\phiout}{\phi^{\text{(out)}}}

\newcommand\nl[1]{``\textit{#1}''}
\newcommand\wl[1]{\texttt{#1}}
\newcommand\smallgap{\vspace{1mm}}

\newcommand{\concsep}{\texttt{</s>}\xspace}
\newcommand{\Gin}{G_\text{in}}
\newcommand{\Gout}{G_\text{out}}

\aclfinalcopy

\title{Data Recombination for Neural Semantic Parsing}

\author{
  Robin Jia \\
  Computer Science Department \\
  Stanford University \\
  {\tt robinjia@stanford.edu}
\And
	Percy Liang \\
  Computer Science Department \\
  Stanford University \\
  {\tt pliang@cs.stanford.edu}
}

\date{}

\begin{document}

\maketitle

\begin{abstract}
Modeling crisp logical regularities is crucial
in semantic parsing,
making it difficult for neural models with no
task-specific prior knowledge to achieve good results.
In this paper, we introduce data recombination,
a novel framework for injecting
such prior knowledge into a model.
From the training data,
we induce a high-precision synchronous context-free grammar,
which captures important conditional independence properties 
commonly found in semantic parsing.
We then train a sequence-to-sequence recurrent network (RNN)
model with a novel attention-based copying mechanism
on datapoints sampled from this grammar,
thereby teaching the model about these structural properties.
Data recombination improves the accuracy
of our RNN model on three semantic parsing datasets, 
leading to new state-of-the-art performance on the standard GeoQuery dataset
for models with comparable supervision.

\end{abstract}

\section{Introduction}
Semantic parsing---the precise translation of natural language utterances into 
logical forms---has many applications,
including question answering
\cite{zelle96geoquery,zettlemoyer05ccg,zettlemoyer07relaxed,liang11dcs,berant2013freebase},
instruction following \cite{artzi2013weakly},
and regular expression generation \cite{kushman2013regex}.
Modern semantic parsers \cite{artzi2013uw,berant2013freebase} 
are complex pieces of software,
requiring hand-crafted features, lexicons, and grammars.

\begin{figure}[t] 
\small
\begin{center} 
  \includegraphics[scale=0.35]{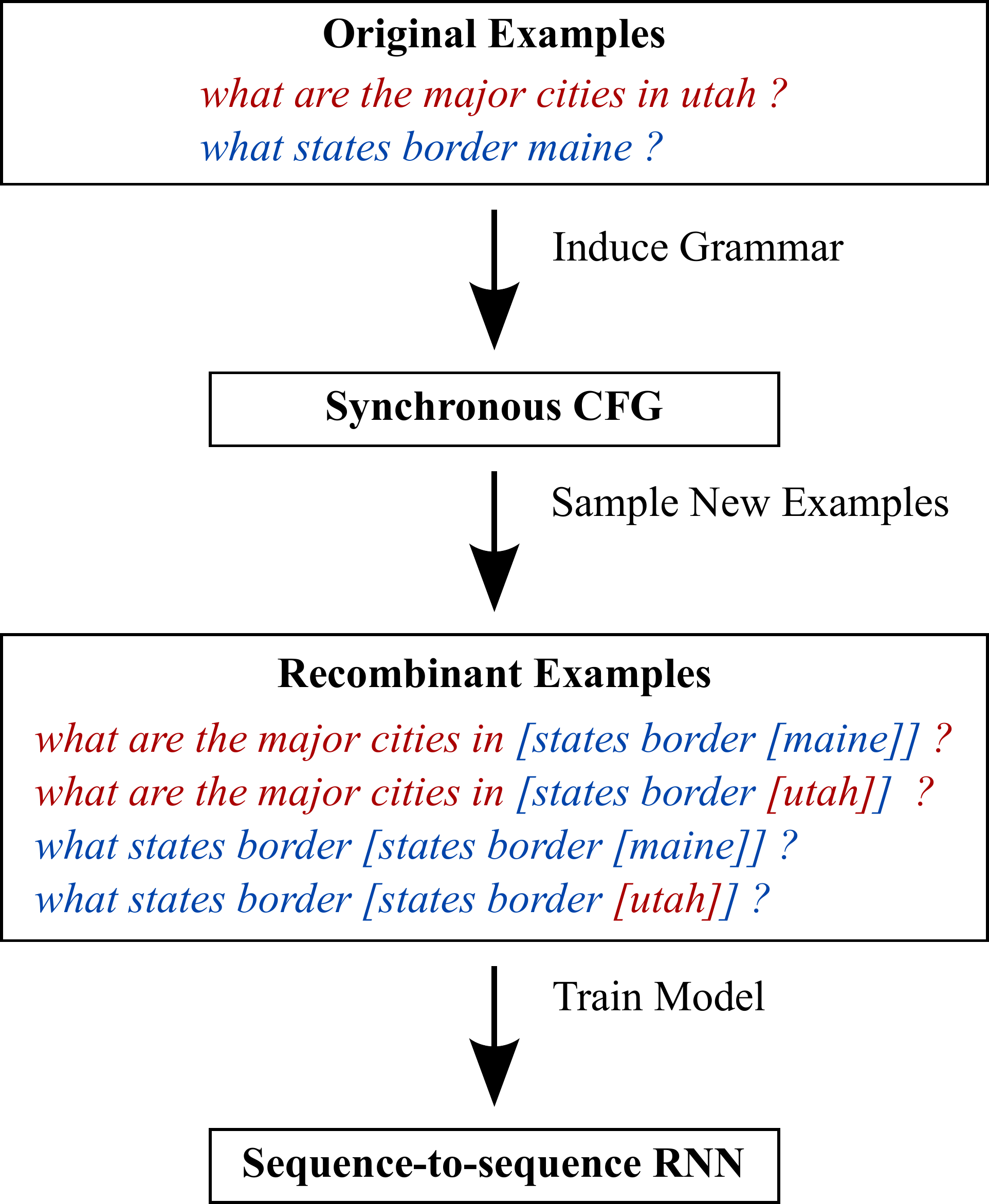}
\end{center} 
\caption{
\label{fig:overview}
  An overview of our system.
  Given a dataset, we induce a high-precision
  synchronous context-free grammar.
  We then sample from this grammar to generate new
  ``recombinant'' examples, which we use to train a sequence-to-sequence RNN.
}
\end{figure}

Meanwhile, recurrent neural networks (RNNs) have made swift inroads into 
many structured prediction tasks in NLP,
including machine translation
\cite{sutskever2014sequence,bahdanau2014neural} and
syntactic parsing \cite{vinyals2015grammar,dyer2015transition}.
Because RNNs make very few domain-specific assumptions,
they have the potential to succeed at a wide variety of tasks
with minimal feature engineering.
However, this flexibility also puts RNNs at a disadvantage compared to
standard semantic parsers, which can generalize naturally
by leveraging their built-in awareness of logical compositionality.

In this paper, we introduce data recombination,
a generic framework for declaratively injecting prior knowledge
into a domain-general structured prediction model.
In data recombination, prior knowledge about a task
is used to build a high-precision generative model
that expands the empirical distribution
by allowing fragments of different examples to be combined in
particular ways.
Samples from this generative model are then used to train a domain-general model.
In the case of semantic parsing, we construct a generative model
by inducing a synchronous context-free grammar (SCFG),
creating new examples such as those shown in Figure~\ref{fig:overview};
our domain-general model is a sequence-to-sequence RNN
with a novel attention-based copying mechanism.
Data recombination boosts the accuracy
of our RNN model on three semantic parsing datasets.
On the \geo dataset, data recombination
improves test accuracy by $4.3$ percentage points
over our baseline RNN, leading to new
state-of-the-art results for models that do not use a
seed lexicon for predicates.

\section{Problem statement}
\begin{figure}[t] 
\small
\begin{framed}
\footnotesize
\textbf{\geo} \\
$x$: \nl{what is the population of iowa ?}\\
$y$: \wl{\_answer ( NV , ( }

\quad \wl{\_population ( NV , V1 ) , \_const (}

\qquad \wl{V0 , \_stateid ( iowa ) ) ) )}

\smallgap
\textbf{\atis}\\
$x$: \nl{can you list all flights from chicago to milwaukee} \\
$y$: \wl{( \_lambda \$0 e ( \_and }

\quad \wl{( \_flight \$0 )}

\quad \wl{( \_from \$0 chicago : \_ci )}

\quad \wl{( \_to \$0 milwaukee : \_ci ) ) )}

\smallgap
\textbf{Overnight} \\
$x$: \nl{when is the weekly standup} \\
$y$: \wl{( call listValue ( call} 

\qquad \wl{getProperty meeting.weekly\_standup}

\qquad \wl{( string start\_time ) ) )}

\end{framed}
\caption{One example from each of our domains.
  We tokenize logical forms as shown, thereby casting 
  semantic parsing as a sequence-to-sequence task.
}
\label{fig:task}
\end{figure}

We cast semantic parsing as a sequence-to-sequence task.
The input utterance $x$ is a sequence of words $x_1, \dotsc, x_m
\in \vocabin$, the input vocabulary;
similarly, the output logical form $y$ is 
a sequence of tokens $y_1, \dotsc, y_n \in \vocabout$, the output vocabulary.
A linear sequence of tokens might appear to 
lose the hierarchical structure of a logical form,
but there is precedent for this choice:
\newcite{vinyals2015grammar}
showed that an RNN can reliably predict tree-structured outputs
in a linear fashion.

We evaluate our system on three existing semantic parsing datasets.
Figure~\ref{fig:task} shows sample input-output pairs from each of these datasets.
\begin{itemize}
  \item \textbf{GeoQuery} (\geo) contains
  natural language questions about US geography
  paired with corresponding Prolog database queries.
  We use the standard split of 600 training examples and 280 test examples
  introduced by \newcite{zettlemoyer05ccg}.
  We preprocess the logical forms to De Brujin index notation
  to standardize variable naming.
  \item \textbf{ATIS} (\atis) contains 
    natural language queries for a flights database
    paired with corresponding database queries
    written in lambda calculus.
    We train on $4473$ examples and evaluate on the $448$
    test examples used by 
    \newcite{zettlemoyer07relaxed}.
  \item \textbf{Overnight} (\overnight) contains
    logical forms paired with natural language
    paraphrases across eight varied subdomains.
    \citet{wang2015overnight} constructed the dataset
    by generating all possible logical
    forms up to some depth threshold, 
    then getting multiple natural language paraphrases
    for each logical form from workers on Amazon Mechanical Turk.
    We evaluate on the same train/test splits as
    \newcite{wang2015overnight}.
\end{itemize}

In this paper, we only explore learning from logical forms.
In the last few years, there has an emergence of
semantic parsers learned from denotations
\cite{clarke10world,liang11dcs,berant2013freebase,artzi2013weakly}.
While our system cannot directly learn from denotations,
it could be used to rerank candidate derivations
generated by one of these other systems.

\section{Sequence-to-sequence RNN Model}
Our sequence-to-sequence RNN model is based on existing 
attention-based neural machine translation models
\cite{bahdanau2014neural,luong2015translation},
but also includes a novel attention-based copying mechanism.
Similar copying mechanisms have been explored
in parallel by \newcite{gu2016copying} and \newcite{gulcehre2016pointing}.

\subsection{Basic Model}
\label{sec:basic}
\paragraph{Encoder.}  The encoder converts the input sequence $x_1, \dotsc, x_m$
into a sequence of \emph{context-sensitive embeddings}
$b_1, \dotsc, b_m$ using a bidirectional RNN \cite{bahdanau2014neural}.
First, a word embedding function $\phiin$ 
maps each word $x_i$ to a fixed-dimensional vector.
These vectors are fed as input to two RNNs: a forward RNN and a backward RNN.
The forward RNN starts with an initial hidden state $h_0^{\text{F}}$,
and generates a sequence of hidden states $h_1^{\text{F}}, \dotsc, h_m^{\text{F}}$ by
repeatedly applying the recurrence 
\begin{align}
  h_i^{\text{F}} = \text{LSTM}(\phiin(x_i), h_{i-1}^{\text{F}}).
\end{align}
The recurrence takes the form of an LSTM \cite{hochreiter1997lstm}.
The backward RNN similarly generates hidden states $h_m^{\text{B}}, \dotsc, h_1^{\text{B}}$
by processing the input sequence in reverse order.
Finally, for each input position $i$, we define 
the context-sensitive embedding
$b_i$ to be the concatenation of $h_i^{\text{F}}$ and $h_i^{\text{B}}$

\paragraph{Decoder.} 
The decoder is an attention-based model
\cite{bahdanau2014neural,luong2015translation}
that generates the output sequence $y_1, \dotsc, y_n$
one token at a time.  At each time step $j$,
it writes $y_{j}$ based on the
current hidden state $s_j$, then updates the hidden
state to $s_{j+1}$ based on $s_j$ and $y_{j}$.
Formally, the decoder is defined by the following equations:
\begin{align}
  s_1 &= \tanh(W^{(s)} [h_m^{\text{F}}, h_1^{\text{B}}]).
  \\ e_{ji} &= s_j^\top W^{(a)} b_i.
  \\ \alpha_{ji} &= \frac{\exp(e_{ji})}{\sum_{i'=1}^m \exp(e_{ji'})}.
  \\ c_j &= \sum_{i=1}^m \alpha_{ji} b_i.
  \\ P(y_{j} &= w \mid x, y_{1:j-1}) \propto \exp(U_w[s_j, c_j]).
  \\ s_{j+1} &= \text{LSTM}([\phiout(y_{j}), c_j], s_j).
\end{align}
When not specified, $i$ ranges over $\{1, \dotsc, m\}$
and $j$ ranges over $\{1, \dotsc, n\}$.
Intuitively, the $\alpha_{ji}$'s define a probability
distribution over the input words,
describing what words in the input the decoder is focusing on at time $j$.
They are computed from the unnormalized
attention scores $e_{ji}$.
The matrices $W^{(s)}$, $W^{(a)}$, and $U$,
as well as the embedding function $\phiout$, are parameters of the model.

\subsection{Attention-based Copying}
In the basic model of the previous section,
the next output word $y_j$ is chosen
via a simple softmax over all words in the output vocabulary.
However, this model has difficulty generalizing to the long tail of
entity names commonly found in semantic parsing datasets.
Conveniently, entity names in the input often correspond 
directly to tokens in the output 
(e.g., \nl{iowa} becomes \wl{iowa} in Figure~\ref{fig:task}).\footnote{
On \geo and \atis, we make a point
not to rely on orthography for non-entities such as \nl{state} to \wl{\_state},
since this leverages information not available to previous models \citep{zettlemoyer05ccg}
and is much less language-independent.}

To capture this intuition, we introduce
a new attention-based copying mechanism.
At each time step $j$, the decoder
generates one of two types of actions.
As before, it can write any word in the output vocabulary.
In addition, it can copy any input word $x_i$ directly to the output,
where the probability with which we copy $x_i$ is determined by the
attention score on $x_i$.
Formally, we define a latent action $a_j$
that is either $\texttt{Write}[w]$ for some $w \in \vocabout$
or $\texttt{Copy}[i]$ for some $i \in \{1, \dotsc, m\}$.
We then have
\begin{align}
  P(a_j &= \texttt{Write}[w]\mid x, y_{1:j-1}) \propto \exp(U_w[s_j, c_j]),
\\ P(a_j &= \texttt{Copy}[i] \mid x, y_{1:j-1}) \propto \exp(e_{ji}).
\end{align}
The decoder chooses $a_j$ with a softmax over all these possible actions;
$y_j$ is then a deterministic function of $a_j$ and $x$.
During training, we maximize the log-likelihood of $y$,
marginalizing out $a$.

Attention-based copying can be seen as a 
combination of a standard softmax output layer of an attention-based model \citep{bahdanau2014neural}
and a Pointer Network \cite{vinyals2015pointer}; in a Pointer Network,
the only way to generate output is to copy a symbol from the input.

\section{Data Recombination}
\begin{figure*}[t] 
\small
\begin{framed}
\footnotesize
\textbf{Examples}

(\nl{what states border texas ?}, \\
\wl{answer(NV, (state(V0), next\_to(V0, NV), const(V0, stateid(texas))))})

(\nl{what is the highest mountain in ohio ?},\\
\wl{answer(NV, highest(V0, (mountain(V0), loc(V0, NV), const(V0, stateid(ohio)))))})

\smallgap
\textbf{Rules created by \absent}

\catroot $\to \langle$ \nl{what states border \catstateid { }?},

\quad \wl{answer(NV, (state(V0), next\_to(V0, NV), const(V0, stateid(}\catstateid\wl{))))}$\rangle$

\catstateid $\to \langle$ \nl{texas}, \wl{texas} $\rangle$

\catroot $\to \langle$ \nl{what is the highest mountain in \catstateid { }?},

\quad \wl{answer(NV, highest(V0, (mountain(V0), loc(V0, NV), }

\qquad \qquad \qquad \qquad \qquad \qquad \enskip \wl{const(V0, stateid(}\catstateid\wl{)))))}$\rangle$

\catstateid $\to \langle$\nl{ohio}, \wl{ohio}$\rangle$

\smallgap
\textbf{Rules created by \absphrase}

\catroot $\to \langle$ \nl{what states border \catstate { }?},
\wl{ answer(NV, (state(V0), next\_to(V0, NV), }\catstate\wl{))}$\rangle$

\catstate $\to \langle$ \nl{states border texas},
\wl{ state(V0), next\_to(V0, NV), const(V0, stateid(texas))}$\rangle$

\catroot $\to \langle$ \nl{what is the highest mountain in \catstate { }?},

\quad \wl{answer(NV, highest(V0, (mountain(V0), loc(V0, NV), }\catstate\wl{)))}$\rangle$

\smallgap
\textbf{Rules created by \concat{2}}

\catroot $\to \langle \catsent_1$ \concsep $\catsent_2, 
\catsent_1$ \concsep $\catsent_2 \rangle$

\catsent $\to \langle$ \nl{what states border texas ?},

\quad \wl{answer(NV, (state(V0), next\_to(V0, NV), const(V0, stateid(texas))))} $\rangle$

\catsent $\to \langle$ \nl{what is the highest mountain in ohio ?},\\
\wl{answer(NV, highest(V0, (mountain(V0), loc(V0, NV), const(V0, stateid(ohio)))))}
$\rangle$

\end{framed}
\caption{Various grammar induction strategies illustrated on \geo.  
  Each strategy converts the rules of an input grammar 
  into rules of an output grammar.
  This figure shows the base case where the
  input grammar has rules
  $\catroot \to \langle x, y \rangle$ 
  for each $(x, y)$ pair in the training dataset.
}
\label{fig:augment-geo}
\end{figure*}
\subsection{Motivation}
The main contribution of this paper is a novel data recombination framework
that injects important prior knowledge into our oblivious sequence-to-sequence RNN.
In this framework, we induce a high-precision
generative model from the training data,
then sample from it to generate new training examples.
The process of inducing this generative model
can leverage any available prior knowledge,
which is transmitted through the generated examples
to the RNN model. 
A key advantage of our two-stage approach is that it allows us to
declare desired properties of the task which might be hard to capture
in the model architecture.

Our approach generalizes data augmentation,
which is commonly employed to inject prior knowledge into a model.
Data augmentation techniques focus on modeling 
invariances---transformations like
translating an image or adding noise
that alter the inputs $x$,
but do not change the output $y$.
These techniques have proven effective in areas like
computer vision \cite{krizhevsky2012imagenet}
and speech recognition \cite{jaitly2013vocal}.

In semantic parsing, however,
we would like to capture more than just invariance properties.
Consider an example with the utterance \nl{what states border texas ?}.
Given this example, it should be easy to 
generalize to questions where \nl{texas}
is replaced by the name of any other state:
simply replace the mention of Texas in the logical form
with the name of the new state.
Underlying this phenomenon is a strong conditional independence principle:
the meaning of the rest of the sentence is independent of the
name of the state in question.
Standard data augmentation is not sufficient to model such phenomena:
instead of holding $y$ fixed, 
we would like to apply simultaneous transformations to $x$ and $y$
such that the new $x$ still maps to the new $y$.
Data recombination addresses this need.

\subsection{General Setting}
In the general setting of data recombination,
we start with a training set $\mathcal{D}$ of $(x, y)$ pairs,
which defines the empirical distribution $\hat p(x, y)$.
We then fit a generative model $\tilde p(x, y)$ to $\hat p$
which generalizes beyond the support of $\hat p$,
for example by splicing together fragments of different examples.
We refer to examples in the support of $\tilde p$ as \emph{recombinant} examples.
Finally, to train our actual model $p_\theta(y \mid x)$, 
we maximize the expected value of
$\log p_\theta(y \mid x)$, where $(x, y)$ is drawn from $\tilde p$.

\subsection{SCFGs for Semantic Parsing}
For semantic parsing, we induce a synchronous context-free grammar (SCFG)
to serve as the backbone of our generative model $\tilde p$.
An SCFG consists of a set of production rules
$X \to \inner{\alpha}{\beta}$, where $X$ is a category (non-terminal),
and $\alpha$ and $\beta$ are sequences of terminal and non-terminal symbols.
Any non-terminal symbols in $\alpha$ must
be aligned to the same non-terminal symbol in $\beta$,
and vice versa.
Therefore, an SCFG defines a set of joint derivations of 
aligned pairs of strings.
In our case, we use an SCFG to represent joint derivations
of utterances $x$ and logical forms $y$ (which for us is just a sequence of tokens). 
After we induce an SCFG $G$ from $\mathcal{D}$,
the corresponding generative model $\tilde p(x, y)$ 
is the distribution over pairs $(x, y)$ 
defined by sampling from $G$,
where we choose production rules to apply uniformly at random.

It is instructive to compare our SCFG-based data recombination with 
\textsc{Wasp} \cite{wong06mt,wong07synchronous},
which uses an SCFG as the actual semantic parsing model.
The grammar induced by \textsc{Wasp}
must have good coverage in order to generalize to new inputs
at test time.
\textsc{Wasp} also requires the implementation of an
efficient algorithm for computing the conditional
probability $p(y \mid x)$.
In contrast, our SCFG is only used to convey
prior knowledge about conditional independence structure,
so it only needs to have high precision;
our RNN model is responsible for boosting recall
over the entire input space.
We also only need to forward sample from the SCFG, which is
considerably easier to implement than conditional inference.

Below, we examine various strategies for inducing 
a grammar $G$ from a dataset $\mathcal D$.
We first encode $\mathcal{D}$ as an initial grammar
with rules \catroot $\to \inner{x}{y}$
for each $(x, y) \in \mathcal{D}$.
Next, we will define each grammar induction strategy
as a mapping from an input grammar $\Gin$ to a new grammar $\Gout$.
This formulation allows us to compose grammar induction strategies 
(Section~\ref{sec:composition}).

\subsubsection{Abstracting Entities}

Our first grammar induction strategy, \absent, simply abstracts entities 
with their types.
We assume that each entity $e$ (e.g., \wl{texas})
has a corresponding type $e.t$ (e.g., \wl{state}),
which we infer based on the presence of certain predicates in the logical form
(e.g. \wl{stateid}).
For each grammar rule $X \to \inner{\alpha}{\beta}$ in $\Gin$,
where $\alpha$ contains a token (e.g., \nl{texas}) that
string matches an entity (e.g., \wl{texas}) in $\beta$,
we add two rules to $\Gout$:
(i) a rule where both occurrences are replaced with the type of the entity 
(e.g., \wl{state}),
and (ii) a new rule that maps the type to the entity (e.g., 
$\catstateid \rightarrow \inner{\text{\nl{texas}}}{\wl{texas}}$; 
we reserve the category name \catstate for the next section).
Thus, $\Gout$ generates recombinant examples
that fuse most of one example with an entity found in a second example.
A concrete example from the \geo domain is given in
Figure~\ref{fig:augment-geo}.

\subsubsection{Abstracting Whole Phrases}

Our second grammar induction strategy, \absphrase, abstracts both entities
and whole phrases with their types.
For each grammar rule $X \to \inner{\alpha}{\beta}$ in $\Gin$,
we add up to two rules to $\Gout$.
First, if $\alpha$ contains tokens
that string match to an entity in $\beta$,
we replace both occurrences with the type of the entity,
similarly to rule (i) from \absent.
Second, if we can infer that the entire expression $\beta$
evaluates to a set of a particular type (e.g. \wl{state})
we create a rule that maps
the type to $\inner{\alpha}{\beta}$.
In practice, we also use some simple rules to strip question identifiers
from $\alpha$, so that the resulting examples are more natural.
Again, refer to Figure~\ref{fig:augment-geo} for a concrete example.

This strategy works because of a more general
conditional independence property:
the meaning of any semantically coherent phrase is 
conditionally independent of the rest of the sentence,
the cornerstone of compositional semantics.
Note that this assumption is not always correct in general:
for example, phenomena like anaphora that involve long-range context 
dependence violate this assumption.  However, this property holds
in most existing semantic parsing datasets.

\subsubsection{Concatenation}

The final grammar induction strategy is a 
surprisingly simple approach we tried that turns out to work.
For any $k \ge 2$, we define the
\concat{$k$} strategy, which creates two types of rules.
First, we create a single rule that has
\catroot going to a sequence of $k$ \catsent's.
Then, for each root-level rule
$\catroot \to \inner{\alpha}{\beta}$ in $\Gin$,
we add the rule $\catsent \to \inner{\alpha}{\beta}$ to $\Gout$.
See Figure~\ref{fig:augment-geo} for an example.

Unlike \absent and \absphrase, 
concatenation is very general, and 
can be applied to any sequence transduction problem.
Of course, it also does not introduce additional information
about compositionality or independence properties present
in semantic parsing. 
However, it does generate harder examples for the attention-based RNN,
since the model must learn to attend to the correct 
parts of the now-longer input sequence.
Related work has shown that training a model on
more difficult examples can improve generalization,
the most canonical case being dropout
\cite{hinton2012improving,wager2013dropout}.

\subsubsection{Composition} \label{sec:composition}
We note that grammar induction strategies
can be composed, yielding more complex grammars.
Given any two grammar induction strategies $f_1$ and $f_2$,
the composition $f_1 \circ f_2$ is the grammar induction strategy
that takes in $\Gin$ and returns $f_1(f_2(\Gin))$.
For the strategies we have defined,
we can perform this operation symbolically on the
grammar rules, without having to sample from the 
intermediate grammar $f_2(\Gin)$.

\section{Experiments}

We evaluate our system on three domains: \geo, \atis, and \overnight.
For \atis, we report logical form exact match accuracy.
For \geo and \overnight, we determine correctness based on denotation match,
as in \newcite{liang11dcs} and \newcite{wang2015overnight}, respectively.

\subsection{Choice of Grammar Induction Strategy}
We note that not all grammar induction strategies
make sense for all domains.
In particular, we only apply \absphrase to \geo and \overnight.
We do not apply \absphrase to \atis, as the dataset
has little nesting structure.

\subsection{Implementation Details}
\begin{figure}[t] 
\small
\begin{framed}
\begin{algorithmic}
  \Function{train}{dataset $D$, number of epochs $T$,
    
  \quad number of examples to sample $n$}
    \State Induce grammar $G$ from $D$
    \State Initialize RNN parameters $\theta$ randomly
    \For{each iteration $t = 1, \dotsc, T$}
      \State Compute current learning rate $\eta_t$
      \State Initialize current dataset $D_t$ to $D$
      \For{$i = 1, \dotsc, n$}
        \State Sample new example $(x', y')$ from $G$
        \State Add $(x', y')$ to $D_t$
      \EndFor
      \State Shuffle $D_t$
      \For{each example $(x, y)$ in $D_t$}
        \State $\theta \gets \theta + \eta_t \nabla \log p_\theta(y \mid x)$
      \EndFor
    \EndFor
  \EndFunction
\end{algorithmic}
\end{framed}
\caption{The training procedure with data recombination.
  We first induce an SCFG, then sample new recombinant examples
  from it at each epoch.}
\label{fig:training}
\end{figure}
We tokenize logical forms in a domain-specific manner,
based on the syntax of the formal language being used.
On \geo and \atis, we disallow copying of predicate names
to ensure a fair comparison to previous work,
as string matching between input words and predicate names
is not commonly used.
We prevent copying by prepending underscores
to predicate tokens; see Figure~\ref{fig:task} for examples.

On \atis alone, when doing attention-based copying and 
data recombination, we leverage an external lexicon
that maps natural language phrases (e.g., \nl{kennedy airport})
to entities (e.g., \wl{jfk:ap}).
When we copy a word that is part of a phrase
in the lexicon, we write the entity 
associated with that lexicon entry.
When performing data recombination,
we identify entity alignments based on 
matching phrases and entities from the lexicon.

We run all experiments with $200$ hidden units
and $100$-dimensional word vectors.
We initialize all parameters uniformly at random 
within the interval $[-0.1, 0.1]$.
We maximize the log-likelihood of the correct logical form using
stochastic gradient descent.
We train the model for a total of $30$ epochs
with an initial learning rate of $0.1$,
and halve the learning rate every $5$ epochs, starting after epoch $15$.
We replace word vectors for words that occur only once in the training set 
with a universal \wl{<unk>} word vector.
Our model is implemented in Theano \cite{bergstra2010theano}.

When performing data recombination, we sample
a new round of recombinant examples from our grammar at each epoch.
We add these examples to the original training dataset,
randomly shuffle all examples, and train the model for the epoch.
Figure~\ref{fig:training} gives pseudocode for this training procedure.
One important hyperparameter is how many examples
to sample at each epoch:
we found that a good rule of thumb is to sample
as many recombinant examples as there are examples
in the training dataset, so that
half of the examples the model sees at each epoch
are recombinant.

At test time, we use beam search with beam size $5$.
We automatically balance missing right parentheses
by adding them at the end. 
On \geo and \overnight, we then pick the highest-scoring logical form
that does not yield an executor error when the
corresponding denotation is computed.
On \atis, we just pick the top prediction on the beam.

\subsection{Impact of the Copying Mechanism}
\begin{table}[t]
  \centering
  \footnotesize
  \begin{tabular}{|l|c|c|c|}
    \hline
    & \geo & \atis & \overnight \\
    \hline
    No Copying   & $74.6$ & $69.9$ & $76.7$ \\
    With Copying & $85.0$ & $76.3$ & $75.8$ \\
    \hline
  \end{tabular}
  \caption{
    Test accuracy on \geo, \atis, and \overnight,
    both with and without copying.  On \overnight,
    we average across all eight domains.}
  \label{tab:copying}
\end{table}
First, we measure the contribution of the
attention-based copying mechanism to the model's overall performance.
On each task, we train and evaluate two models: one with the copying mechanism,
and one without.  Training is done without data recombination.
The results are shown in Table~\ref{tab:copying}.

On \geo and \atis, the copying mechanism
helps significantly: it improves test accuracy by
$10.4$ percentage points on \geo and $6.4$ points on \atis.
However, on \overnight, adding the copying mechanism
actually makes our model perform slightly worse.
This result is somewhat expected, as the \overnight 
dataset contains a very small number of distinct entities.
It is also notable that both systems surpass the
previous best system on \overnight by a wide margin.

We choose to use the copying mechanism in all subsequent experiments,
as it has a large advantage in realistic settings
where there are many distinct entities in the world.
The concurrent work of 
\newcite{gu2016copying} and \newcite{gulcehre2016pointing},
both of whom propose similar copying mechanisms,
provides additional evidence for the utility of copying
on a wide range of NLP tasks.

\subsection{Main Results}
\begin{table}[t]
  \centering
  \footnotesize
  \begin{tabular}{|l|c|c|c|}
    \hline
    & \geo & \atis \\
    \hline
    \textbf{Previous Work} & & \\
    \newcite{zettlemoyer07relaxed} &        & $\bf 84.6$ \\
    \newcite{kwiatkowski10ccg}     & $88.9$ &        \\
    \newcite{liang11dcs}\footnotemark
      & $91.1$ &        \\
    \newcite{kwiatkowski11lex}     & $88.6$ & $82.8$ \\
    \newcite{poon2013gusp}         &        & $83.5$ \\
    \newcite{zhao2015type}         & $88.9$ & $84.2$ \\
    \hline
    \textbf{Our Model} & & \\
    No Recombination               & $85.0$ & $76.3$ \\
    \absent                        & $85.4$ & $79.9$ \\
    \absphrase                     & $87.5$ &        \\
    \concat{2}                     & $84.6$ & $79.0$ \\
    \concat{3}                     &        & $77.5$ \\
    AWP + AE                       & $88.9$ &        \\
    AE + C2                        &        & $78.8$ \\
    AWP + AE + C2                  & $\bf 89.3$ &        \\
    AE + C3                        &        & $83.3$ \\
    \hline
  \end{tabular}
  \caption{Test accuracy using different data recombination strategies
    on \geo and \atis.
    AE is \absent, AWP is \absphrase, C2 is \concat{2}, and C3 is \concat{3}.
  }
  \label{tab:results}
\end{table}
\begin{table*}[t]
  \centering
  \scalebox{0.68}{
  \begin{tabular}{|l|cccccccc|c|}
    \hline
    & \textsc{Basketball} & \textsc{Blocks} & \textsc{Calendar} 
    & \textsc{Housing} & \textsc{Publications} & \textsc{Recipes} 
    & \textsc{Restaurants} & \textsc{Social} & Avg. \\
    \hline
    \textbf{Previous Work} & & & & & & & & & \\
    \newcite{wang2015overnight}
                      & $46.3$ & $41.9$ & $74.4$ & $54.0$ & $59.0$ & $70.8$ & $75.9$ & $48.2$ & $58.8$\\
    \hline
    \textbf{Our Model} & & & & & & & & & \\
    No Recombination  & $85.2$ & $58.1$ & $78.0$ & $71.4$ & $76.4$ & $79.6$ & $76.2$ & $81.4$ & $75.8$\\
    \absent           & $86.7$ & $60.2$ & $78.0$ & $65.6$ & $73.9$ & $77.3$ & $79.5$ & $81.3$ & $75.3$\\
    \absphrase        & $86.7$ & $55.9$ & $79.2$ & $69.8$ & $76.4$ & $77.8$ & $80.7$ & $80.9$ & $75.9$\\
    \concat{2}        & $84.7$ & $\bf 60.7$ & $75.6$ & $69.8$ & $74.5$ & $80.1$ & $79.5$ & $80.8$ & $75.7$\\
    AWP + AE          & $85.2$ & $54.1$ & $78.6$ & $67.2$ & $73.9$ & $79.6$ & $\bf 81.9$ & $\bf 82.1$ & $75.3$\\
    AWP + AE + C2     & $\bf 87.5$ & $60.2$ & $\bf 81.0$ & $\bf 72.5$ & $\bf 78.3$ & $\bf 81.0$ & $79.5$ & $79.6$ & $\bf 77.5$\\
    \hline
  \end{tabular}}
  \caption{Test accuracy using different data recombination strategies on the \overnight tasks.}
  \label{tab:overnight}
\end{table*}

\footnotetext{The method of \newcite{liang11dcs}
is not comparable to ours, as they
as they used a seed lexicon mapping words to predicates.
We explicitly avoid using such prior knowledge in our system.}

For our main results, we train our model with 
a variety of data recombination strategies on all three datasets.
These results are summarized in Tables~\ref{tab:results} and \ref{tab:overnight}.
We compare our system to the baseline of not using any data recombination,
as well as to state-of-the-art systems on all three datasets.

We find that data recombination consistently improves
accuracy across the three domains we evaluated on,
and that the strongest results come from composing multiple strategies.
Combining \absphrase, \absent, and \concat{2}
yields a $4.3$ percentage point improvement over the baseline 
without data recombination on \geo,
and an average of $1.7$ percentage points on \overnight.
In fact, on \geo, we achieve test accuracy of $89.3\%$,
which surpasses the previous state-of-the-art,
excluding \citet{liang11dcs}, which used a seed lexicon for predicates.
On \atis, we experiment with concatenating more than $2$
examples, to make up for the fact that we cannot apply
\absphrase, which generates longer examples.
We obtain a test accuracy of $83.3$ with \absent composed with \concat{3},
which beats the baseline by $7$ percentage points
and is competitive with the state-of-the-art.

\paragraph{Data recombination without copying.}
For completeness, we also investigated the effects of 
data recombination on the model without attention-based copying.
We found that recombination helped significantly on \geo and \atis,
but hurt the model slightly on \overnight.
On \geo, the best data recombination strategy
yielded test accuracy of $82.9\%$, for a gain of $8.3$ percentage
points over the baseline with no copying and no recombination;
on \atis, data recombination gives test accuracies
as high as $74.6\%$, a $4.7$ point gain over the same baseline.
However, no data recombination strategy improved average test accuracy 
on \overnight; the best one resulted in a $0.3$ percentage point
decrease in test accuracy.
We hypothesize that data recombination helps less on
\overnight in general because the space of possible 
logical forms is very limited, making it more like a 
large multiclass classification task.
Therefore, it is less important for the model to learn
good compositional representations that generalize to new 
logical forms at test time.

\subsection{Effect of Longer Examples}
\begin{figure}[t] 
\small
\begin{framed}
\footnotesize
\textbf{Depth-2 (same length)}

$x$: \nl{rel:12 of rel:17 of ent:14}

$y$: \wl{( \_rel:12 ( \_rel:17 \_ent:14 ) )}

\smallgap
\textbf{Depth-4 (longer)}

$x$: \nl{rel:23 of rel:36 of rel:38 of rel:10 of ent:05}

$y$: \wl{( \_rel:23 ( \_rel:36 ( \_rel:38}

\qquad \qquad \wl{( \_rel:10 \_ent:05 ) ) ) )}

\end{framed}
\caption{A sample of our artificial data.}
\label{fig:artificial-data}
\end{figure}

\begin{figure}[t] 
\small
\begin{center} 
  \vspace{-0.35in}
  \includegraphics[scale=0.65]{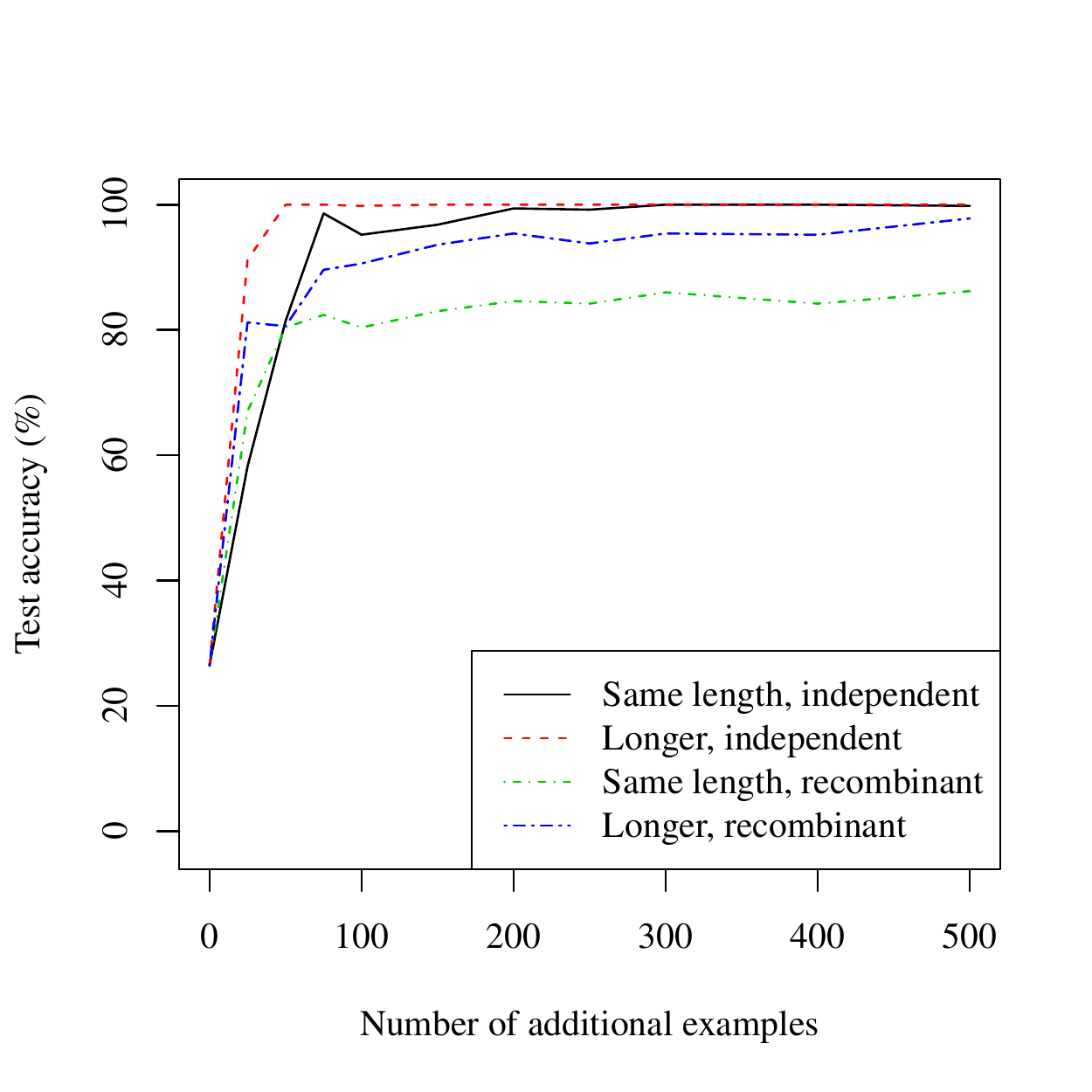}
\end{center} 
  \vspace{-0.15in}
\caption{The results of our artificial data experiments.
  We see that the model learns more from
  longer examples than from same-length examples.
}
\label{fig:artificial}
\end{figure}
Interestingly, strategies like \absphrase and \concat{2}
help the model even though the resulting recombinant examples
are generally not in the support of the test distribution.
In particular, these recombinant examples are on average 
longer than those in the actual dataset,
which makes them harder for the attention-based model.
Indeed, for every domain, our best accuracy numbers 
involved some form of concatenation, and often involved
\absphrase as well.  In comparison, applying \absent
alone, which generates examples of the same length as
those in the original dataset, was generally less effective.

We conducted additional experiments on artificial data
to investigate the importance of adding longer, harder examples.
We experimented with adding new examples via data recombination,
as well as adding new independent examples (e.g. to simulate
the acquisition of more training data).
We constructed a simple world containing a set of entities
and a set of binary relations.
For any $n$, we can generate a set of depth-$n$ examples,
which involve the composition of $n$ relations
applied to a single entity.
Example data points are shown in Figure~\ref{fig:artificial-data}.
We train our model on various datasets,
then test it on a set of $500$ randomly chosen depth-$2$ examples.
The model always has access to a small seed training set of $100$ depth-$2$ examples.
We then add one of four types of examples to the training set:
\begin{itemize}
  \setlength\itemsep{0pt}
  \item \textbf{Same length, independent}: New randomly chosen depth-$2$ examples.\footnote{Technically, these are not completely independent, as we
    sample these new examples without replacement.  The same applies to the
  longer ``independent'' examples.}
  \item \textbf{Longer, independent}: Randomly chosen depth-$4$ examples.
  \item \textbf{Same length, recombinant}: 
    Depth-$2$ examples sampled from
    the grammar induced by applying \absent to the seed dataset.
  \item \textbf{Longer, recombinant}: 
    Depth-$4$ examples sampled from the grammar
    induced by applying \absphrase followed by \absent to the seed dataset.
\end{itemize}
To maintain consistency between the independent and recombinant
experiments, we fix the recombinant examples across all epochs,
instead of resampling at every epoch.
In Figure~\ref{fig:artificial}, we plot accuracy on the test set 
versus the number of additional examples added of each of these four types.
As expected, independent examples are more helpful
than the recombinant ones, but both help the model improve considerably.
In addition, we see that even though the test dataset only has short examples,
adding longer examples helps the model
more than adding shorter ones,
in both the independent and recombinant cases.
These results underscore the importance 
training on longer, harder examples.

\section{Discussion}
\label{sec:discussion}

In this paper, we have presented a novel framework
we term data recombination,
in which we generate new training examples from a 
high-precision generative model
induced from the original training dataset.
We have demonstrated its effectiveness in improving the
accuracy of a sequence-to-sequence RNN model
on three semantic parsing datasets,
using a synchronous context-free grammar
as our generative model.

There has been growing interest in applying neural networks
to semantic parsing and related tasks.
\newcite{dong2016logical} concurrently 
developed an attention-based RNN model for 
semantic parsing, although they did not use data recombination.
\newcite{grefenstette2014deep}
proposed a non-recurrent neural model for semantic parsing,
though they did not run experiments.
\newcite{mei2016listen} use an RNN model
to perform a related task of instruction following.

Our proposed attention-based copying mechanism bears a strong resemblance
to two models that were developed independently by other groups.
\newcite{gu2016copying} apply a very similar copying mechanism
to text summarization and single-turn dialogue generation.
\newcite{gulcehre2016pointing} propose a model that decides at each step
whether to write from a ``shortlist'' vocabulary or copy from the input,
and report improvements on machine translation and text summarization.
Another piece of related work is \newcite{luong2015rare},
who train a neural machine translation system to copy rare words,
relying on an external system to generate alignments.

Prior work has explored using paraphrasing
for data augmentation on NLP tasks.
\newcite{zhang2015character} augment their data
by swapping out words for synonyms from WordNet.
\newcite{wang2015petpeeves} use a similar strategy,
but identify similar words and phrases based on
cosine distance between vector space embeddings.
Unlike our data recombination strategies,
these techniques only change inputs $x$, while keeping the labels $y$ fixed.
Additionally, these paraphrasing-based transformations can be described
in terms of grammar induction, so they can be
incorporated into our framework.

In data recombination, data generated by a high-precision generative model
is used to train a second, domain-general model.
Generative oversampling \cite{liu2007oversampling}
learns a generative model in a multiclass classification setting,
then uses it to generate additional examples from rare classes
in order to combat label imbalance.
Uptraining \cite{petrov2010uptraining} 
uses data labeled by an accurate but slow model to train
a computationally cheaper second model.
\newcite{vinyals2015grammar} generate a large dataset
of constituency parse trees by taking sentences
that multiple existing systems parse in the same way,
and train a neural model on this dataset.

Some of our induced grammars
generate examples that are not in the test distribution,
but nonetheless aid in generalization.
Related work has also explored the idea of training on 
altered or out-of-domain data, often interpreting
it as a form of regularization.
Dropout training has been shown to be a form of adaptive regularization 
\cite{hinton2012improving,wager2013dropout}.
\newcite{guu2015traversing} 
showed that encouraging a knowledge base completion model
to handle longer path queries 
acts as a form of structural regularization.

Language is a blend of crisp regularities and soft relationships.
Our work takes RNNs, which excel at modeling soft phenomena,
and uses a highly structured tool---synchronous context free grammars---to
infuse them with an understanding of crisp structure.
We believe this paradigm for 
simultaneously modeling the soft and hard aspects of language
should have broader applicability beyond semantic parsing.

\paragraph{Acknowledgments}
This work was supported by the 
NSF Graduate Research Fellowship under Grant No. DGE-114747,
and the DARPA Communicating with Computers (CwC) program under ARO
prime contract no. W911NF-15-1-0462.

\paragraph{Reproducibility.} All code, data, and experiments for this
paper are available on the CodaLab platform at
{\small \url{https://worksheets.codalab.org/worksheets/0x50757a37779b485f89012e4ba03b6f4f/}}.

\bibliographystyle{acl2016}
\bibliography{refdb/all}

\begin{thebibliography}{}

\bibitem[\protect\citename{Artzi and Zettlemoyer}2013a]{artzi2013uw}
Y.~Artzi and L.~Zettlemoyer.
\newblock 2013a.
\newblock {UW} {SPF}: The {U}niversity of {W}ashington semantic parsing
  framework.
\newblock {\em arXiv preprint arXiv:1311.3011}.

\bibitem[\protect\citename{Artzi and Zettlemoyer}2013b]{artzi2013weakly}
Y.~Artzi and L.~Zettlemoyer.
\newblock 2013b.
\newblock Weakly supervised learning of semantic parsers for mapping
  instructions to actions.
\newblock {\em Transactions of the Association for Computational Linguistics
  (TACL)}, 1:49--62.

\bibitem[\protect\citename{Bahdanau \bgroup et al.\egroup
  }2014]{bahdanau2014neural}
D.~Bahdanau, K.~Cho, and Y.~Bengio.
\newblock 2014.
\newblock Neural machine translation by jointly learning to align and
  translate.
\newblock {\em arXiv preprint arXiv:1409.0473}.

\bibitem[\protect\citename{Berant \bgroup et al.\egroup
  }2013]{berant2013freebase}
J.~Berant, A.~Chou, R.~Frostig, and P.~Liang.
\newblock 2013.
\newblock Semantic parsing on {F}reebase from question-answer pairs.
\newblock In {\em Empirical Methods in Natural Language Processing (EMNLP)}.

\bibitem[\protect\citename{Bergstra \bgroup et al.\egroup
  }2010]{bergstra2010theano}
J.~Bergstra, O.~Breuleux, F.~Bastien, P.~Lamblin, R.~Pascanu, G.~Desjardins,
  J.~Turian, D.~Warde-Farley, and Y.~Bengio.
\newblock 2010.
\newblock Theano: a {CPU} and {GPU} math expression compiler.
\newblock In {\em Python for Scientific Computing Conference}.

\bibitem[\protect\citename{Clarke \bgroup et al.\egroup }2010]{clarke10world}
J.~Clarke, D.~Goldwasser, M.~Chang, and D.~Roth.
\newblock 2010.
\newblock Driving semantic parsing from the world's response.
\newblock In {\em Computational Natural Language Learning (CoNLL)}, pages
  18--27.

\bibitem[\protect\citename{Dong and Lapata}2016]{dong2016logical}
L.~Dong and M.~Lapata.
\newblock 2016.
\newblock Language to logical form with neural attention.
\newblock In {\em Association for Computational Linguistics (ACL)}.

\bibitem[\protect\citename{Dyer \bgroup et al.\egroup
  }2015]{dyer2015transition}
C.~Dyer, M.~Ballesteros, W.~Ling, A.~Matthews, and N.~A. Smith.
\newblock 2015.
\newblock Transition-based dependency parsing with stack long short-term
  memory.
\newblock In {\em Association for Computational Linguistics (ACL)}.

\bibitem[\protect\citename{Grefenstette \bgroup et al.\egroup
  }2014]{grefenstette2014deep}
E.~Grefenstette, P.~Blunsom, N.~de~Freitas, and K.~M. Hermann.
\newblock 2014.
\newblock A deep architecture for semantic parsing.
\newblock In {\em ACL Workshop on Semantic Parsing}, pages 22--27.

\bibitem[\protect\citename{Gu \bgroup et al.\egroup }2016]{gu2016copying}
J.~Gu, Z.~Lu, H.~Li, and V.~O. Li.
\newblock 2016.
\newblock Incorporating copying mechanism in sequence-to-sequence learning.
\newblock In {\em Association for Computational Linguistics (ACL)}.

\bibitem[\protect\citename{Gulcehre \bgroup et al.\egroup
  }2016]{gulcehre2016pointing}
C.~Gulcehre, S.~Ahn, R.~Nallapati, B.~Zhou, and Y.~Bengio.
\newblock 2016.
\newblock Pointing the unknown words.
\newblock In {\em Association for Computational Linguistics (ACL)}.

\bibitem[\protect\citename{Guu \bgroup et al.\egroup }2015]{guu2015traversing}
K.~Guu, J.~Miller, and P.~Liang.
\newblock 2015.
\newblock Traversing knowledge graphs in vector space.
\newblock In {\em Empirical Methods in Natural Language Processing (EMNLP)}.

\bibitem[\protect\citename{Hinton \bgroup et al.\egroup
  }2012]{hinton2012improving}
G.~E. Hinton, N.~Srivastava, A.~Krizhevsky, I.~Sutskever, and R.~R.
  Salakhutdinov.
\newblock 2012.
\newblock Improving neural networks by preventing co-adaptation of feature
  detectors.
\newblock {\em arXiv preprint arXiv:1207.0580}.

\bibitem[\protect\citename{Hochreiter and Schmidhuber}1997]{hochreiter1997lstm}
S.~Hochreiter and J.~Schmidhuber.
\newblock 1997.
\newblock Long short-term memory.
\newblock {\em Neural Computation}, 9(8):1735--1780.

\bibitem[\protect\citename{Jaitly and Hinton}2013]{jaitly2013vocal}
N.~Jaitly and G.~E. Hinton.
\newblock 2013.
\newblock Vocal tract length perturbation (vtlp) improves {s}peech recognition.
\newblock In {\em International Conference on Machine Learning (ICML)}.

\bibitem[\protect\citename{Krizhevsky \bgroup et al.\egroup
  }2012]{krizhevsky2012imagenet}
A.~Krizhevsky, I.~Sutskever, and G.~E. Hinton.
\newblock 2012.
\newblock Imagenet classification with deep convolutional neural networks.
\newblock In {\em Advances in Neural Information Processing Systems (NIPS)},
  pages 1097--1105.

\bibitem[\protect\citename{Kushman and Barzilay}2013]{kushman2013regex}
N.~Kushman and R.~Barzilay.
\newblock 2013.
\newblock Using semantic unification to generate regular expressions from
  natural language.
\newblock In {\em Human Language Technology and North American Association for
  Computational Linguistics (HLT/NAACL)}, pages 826--836.

\bibitem[\protect\citename{Kwiatkowski \bgroup et al.\egroup
  }2010]{kwiatkowski10ccg}
T.~Kwiatkowski, L.~Zettlemoyer, S.~Goldwater, and M.~Steedman.
\newblock 2010.
\newblock Inducing probabilistic {CCG} grammars from logical form with
  higher-order unification.
\newblock In {\em Empirical Methods in Natural Language Processing (EMNLP)},
  pages 1223--1233.

\bibitem[\protect\citename{Kwiatkowski \bgroup et al.\egroup
  }2011]{kwiatkowski11lex}
T.~Kwiatkowski, L.~Zettlemoyer, S.~Goldwater, and M.~Steedman.
\newblock 2011.
\newblock Lexical generalization in {CCG} grammar induction for semantic
  parsing.
\newblock In {\em Empirical Methods in Natural Language Processing (EMNLP)},
  pages 1512--1523.

\bibitem[\protect\citename{Liang \bgroup et al.\egroup }2011]{liang11dcs}
P.~Liang, M.~I. Jordan, and D.~Klein.
\newblock 2011.
\newblock Learning dependency-based compositional semantics.
\newblock In {\em Association for Computational Linguistics (ACL)}, pages
  590--599.

\bibitem[\protect\citename{Liu \bgroup et al.\egroup
  }2007]{liu2007oversampling}
A.~Liu, J.~Ghosh, and C.~Martin.
\newblock 2007.
\newblock Generative oversampling for mining imbalanced datasets.
\newblock In {\em International Conference on Data Mining (DMIN)}.

\bibitem[\protect\citename{Luong \bgroup et al.\egroup
  }2015a]{luong2015translation}
M.~Luong, H.~Pham, and C.~D. Manning.
\newblock 2015a.
\newblock Effective approaches to attention-based neural machine translation.
\newblock In {\em Empirical Methods in Natural Language Processing (EMNLP)},
  pages 1412--1421.

\bibitem[\protect\citename{Luong \bgroup et al.\egroup }2015b]{luong2015rare}
M.~Luong, I.~Sutskever, Q.~V. Le, O.~Vinyals, and W.~Zaremba.
\newblock 2015b.
\newblock Addressing the rare word problem in neural machine translation.
\newblock In {\em Association for Computational Linguistics (ACL)}, pages
  11--19.

\bibitem[\protect\citename{Mei \bgroup et al.\egroup }2016]{mei2016listen}
H.~Mei, M.~Bansal, and M.~R. Walter.
\newblock 2016.
\newblock Listen, attend, and walk: Neural mapping of navigational instructions
  to action sequences.
\newblock In {\em Association for the Advancement of Artificial Intelligence
  (AAAI)}.

\bibitem[\protect\citename{Petrov \bgroup et al.\egroup
  }2010]{petrov2010uptraining}
S.~Petrov, P.~Chang, M.~Ringgaard, and H.~Alshawi.
\newblock 2010.
\newblock Uptraining for accurate deterministic question parsing.
\newblock In {\em Empirical Methods in Natural Language Processing (EMNLP)}.

\bibitem[\protect\citename{Poon}2013]{poon2013gusp}
H.~Poon.
\newblock 2013.
\newblock Grounded unsupervised semantic parsing.
\newblock In {\em Association for Computational Linguistics (ACL)}.

\bibitem[\protect\citename{Sutskever \bgroup et al.\egroup
  }2014]{sutskever2014sequence}
I.~Sutskever, O.~Vinyals, and Q.~V. Le.
\newblock 2014.
\newblock Sequence to sequence learning with neural networks.
\newblock In {\em Advances in Neural Information Processing Systems (NIPS)},
  pages 3104--3112.

\bibitem[\protect\citename{Vinyals \bgroup et al.\egroup
  }2015a]{vinyals2015pointer}
O.~Vinyals, M.~Fortunato, and N.~Jaitly.
\newblock 2015a.
\newblock Pointer networks.
\newblock In {\em Advances in Neural Information Processing Systems (NIPS)},
  pages 2674--2682.

\bibitem[\protect\citename{Vinyals \bgroup et al.\egroup
  }2015b]{vinyals2015grammar}
O.~Vinyals, L.~Kaiser, T.~Koo, S.~Petrov, I.~Sutskever, and G.~Hinton.
\newblock 2015b.
\newblock Grammar as a foreign language.
\newblock In {\em Advances in Neural Information Processing Systems (NIPS)},
  pages 2755--2763.

\bibitem[\protect\citename{Wager \bgroup et al.\egroup }2013]{wager2013dropout}
S.~Wager, S.~I. Wang, and P.~Liang.
\newblock 2013.
\newblock Dropout training as adaptive regularization.
\newblock In {\em Advances in Neural Information Processing Systems (NIPS)}.

\bibitem[\protect\citename{Wang and Yang}2015]{wang2015petpeeves}
W.~Y. Wang and D.~Yang.
\newblock 2015.
\newblock That’s so annoying!!!: A lexical and frame-semantic embedding based
  data augmentation approach to automatic categorization of annoying behaviors
  using \#petpeeve tweets.
\newblock In {\em Empirical Methods in Natural Language Processing (EMNLP)}.

\bibitem[\protect\citename{Wang \bgroup et al.\egroup }2015]{wang2015overnight}
Y.~Wang, J.~Berant, and P.~Liang.
\newblock 2015.
\newblock Building a semantic parser overnight.
\newblock In {\em Association for Computational Linguistics (ACL)}.

\bibitem[\protect\citename{Wong and Mooney}2006]{wong06mt}
Y.~W. Wong and R.~J. Mooney.
\newblock 2006.
\newblock Learning for semantic parsing with statistical machine translation.
\newblock In {\em North American Association for Computational Linguistics
  (NAACL)}, pages 439--446.

\bibitem[\protect\citename{Wong and Mooney}2007]{wong07synchronous}
Y.~W. Wong and R.~J. Mooney.
\newblock 2007.
\newblock Learning synchronous grammars for semantic parsing with lambda
  calculus.
\newblock In {\em Association for Computational Linguistics (ACL)}, pages
  960--967.

\bibitem[\protect\citename{Zelle and Mooney}1996]{zelle96geoquery}
M.~Zelle and R.~J. Mooney.
\newblock 1996.
\newblock Learning to parse database queries using inductive logic programming.
\newblock In {\em Association for the Advancement of Artificial Intelligence
  (AAAI)}, pages 1050--1055.

\bibitem[\protect\citename{Zettlemoyer and Collins}2005]{zettlemoyer05ccg}
L.~S. Zettlemoyer and M.~Collins.
\newblock 2005.
\newblock Learning to map sentences to logical form: Structured classification
  with probabilistic categorial grammars.
\newblock In {\em Uncertainty in Artificial Intelligence (UAI)}, pages
  658--666.

\bibitem[\protect\citename{Zettlemoyer and Collins}2007]{zettlemoyer07relaxed}
L.~S. Zettlemoyer and M.~Collins.
\newblock 2007.
\newblock Online learning of relaxed {CCG} grammars for parsing to logical
  form.
\newblock In {\em Empirical Methods in Natural Language Processing and
  Computational Natural Language Learning (EMNLP/CoNLL)}, pages 678--687.

\bibitem[\protect\citename{Zhang \bgroup et al.\egroup
  }2015]{zhang2015character}
X.~Zhang, J.~Zhao, and Y.~LeCun.
\newblock 2015.
\newblock Character-level convolutional networks for text classification.
\newblock In {\em Advances in Neural Information Processing Systems (NIPS)}.

\bibitem[\protect\citename{Zhao and Huang}2015]{zhao2015type}
K.~Zhao and L.~Huang.
\newblock 2015.
\newblock Type-driven incremental semantic parsing with polymorphism.
\newblock In {\em North American Association for Computational Linguistics
  (NAACL)}.

\end{thebibliography}

\end{document}